\documentclass[english]{sbrt}
\usepackage[english]{babel}
\usepackage[utf8]{inputenc}

\usepackage{url}

\usepackage{graphicx}%
\usepackage{multirow}%
\usepackage{amsmath,amssymb,amsfonts}%
\usepackage{mathtools}
\usepackage{mathrsfs}%
\usepackage{xcolor}%
\usepackage{textcomp}%
\usepackage{manyfoot}%
\usepackage{booktabs}%
\usepackage{algorithm}%
\usepackage{algorithmicx}%
\usepackage{algpseudocode}%
\usepackage{listings}%
\usepackage[acronym]{glossaries}
\usepackage{caption}
\usepackage{subcaption}
\usepackage{cite}

\usepackage{booktabs}

\usepackage{flushend}

\usepackage[caption=false,font=footnotesize]{subfig}
\begin{document}

\title{Gromov–Wasserstein Methods for Multi-View Relational Embedding and Clustering}

\author{Rafael Pereira Eufrazio, Eduardo Fernandes Montesuma, and Charles Casimiro Cavalcante
\thanks{R. P. Eufrazio is with the Instituto Federal de Educação, Ciência e Tecnologia do Ceará, Canindé, Ceará, Brazil,
and also with the Federal University of Ceara, Fortaleza, Ceará, Brazil, e-mail: rafael.eufrazio@ifce.edu.br.}%
\thanks{E. F. Montesuma is with Sigma Nova, Paris, France, e-mail: eduardo.montesuma@sigmanova.ai.}%
\thanks{C. C. Cavalcante is with the Federal University of Ceara, Fortaleza, Ceará, Brazil, e-mail: charles@ufc.br.}%
}

\maketitle


\begin{abstract}
Learning low-dimensional representations from multi-view relational data is challenging when underlying geometries differ across views. We propose Bary-GWMDS, a Gromov-Wasserstein-based method that operates directly on distance matrices to learn a consensus embedding preserving shared relational structure. By leveraging intrinsic distances, the approach naturally handles nonlinear distortions across views. We also introduce Mean-GWMDS-C, a clustering-oriented formulation that averages distance matrices and learns reduced-support representations via a consensus Gromov-Wasserstein transport. Experiments on synthetic and real-world datasets show that the proposed framework yields stable and geometrically meaningful embeddings.
\end{abstract}

\begin{keywords}
Gromov-Wasserstein distance, manifold learning, multi-view learning, relational embedding, semi-relaxed Gromov-Wasserstein
\end{keywords}

\section{Introduction}

Learning low-dimensional representations from relational data
is a fundamental problem in signal processing and machine learning, with applications ranging from manifold learning to multi-modal data analysis \cite{yu2025review}. In signal processing applications, multi-view data naturally arise from heterogeneous sensing modalities, such as sensor networks, image descriptors, and time-series measurements, where each view captures complementary aspects of the underlying signal structure. In many practical scenarios, the same set of samples is observed through multiple
views, each inducing a distinct relational structure \cite{yu2025review,qin2025survey}. Effectively aggregating such heterogeneous view-dependent information into a single embedding remains challenging, especially when the underlying geometries differ across views.

Classical multi-view embedding methods typically rely on Euclidean averaging or linear alignment strategies, which may fail under nonlinear distortions or heterogeneous geometric structures \cite{xu2013survey, chowdhury2025deep, salter2017latent, lin2023multi}. In contrast, the Gromov-Wasserstein (GW) distance provides an intrinsic notion of similarity between metric spaces, enabling the comparison and aggregation of relational data independently of the ambient representation \cite{kerdoncuff2021sampled, montesuma2023recent}. This makes GW-based approaches particularly well suited for multi-view settings where rigid alignments or pointwise correspondences are inadequate.

In this work, we propose Bary-GWMDS, a multi-view embedding method based on the GW barycenter. The approach aggregates view-specific distance matrices into a consensus relational structure and computes a low-dimensional embedding via Gromov-Wasserstein multidimensional scaling. By operating directly on relational information, Bary-GWMDS naturally accommodates heterogeneous views. We also introduce a clustering-oriented formulation, termed Mean-GWMDS-C, building upon the Mean-GWMDS framework originally proposed in \cite{eufrazio2026structure}, which combines view-wise distance averaging with reduced-support embeddings for clustering analysis.

This paper proposes a Gromov-Wasserstein-based framework for multi-view relational learning, comprising a barycentric embedding formulation and a reduced-support clustering-oriented variant. The main contributions are:
\begin{itemize}
\item A GW-based multi-view embedding method using a barycentric formulation;
\item A reduced-support clustering method.
\end{itemize}

Experimental results on synthetic and real-world datasets demonstrate the effectiveness of the proposed approach.



The remainder of this paper is organized as follows. Section II introduces the background, including the problem setup and the Gromov-Wasserstein formulation. Section III presents the proposed multi-view framework, detailing both the Bary-GWMDS and Mean-GWMDS-C approaches. Section IV describes the experimental setup and discusses the results on synthetic and real-world multi-view datasets. Finally, Section V concludes the paper.

\section{Background}

In this section, we present the necessary background on Gromov-Wasserstein geometry and multi-view relational modeling, starting with the problem formulation.

\subsection{Problem Setup}
We consider a multi-view dataset consisting of $S$ views defined over a common
set of $n$ samples. Each view is represented by a relational metric space $(\mathcal{X}, D^{(s)})$, where $D^{(s)} \in \mathbb{R}^{n \times n}$ denotes the pairwise distance matrix associated with view $s$. 

Our goal is to learn a low-dimensional embedding that captures a consensus relational structure across views, based solely on their pairwise distance representations. Such representations can also be leveraged for downstream tasks such as clustering.

\subsection{Gromov-Wasserstein Framework}

The Gromov-Wasserstein (GW) distance compares probability distributions defined on different metric spaces by aligning their intrinsic relational structures \cite{Eufrazio}. 

Given two metric measure spaces $(\mathcal{X}, D_X, \hat{\mu})$ and $(\mathcal{Y}, D_Y, \hat{\nu})$, and denoting $D_X^{ik} = D_X(x_i, x_k)$ and $D_Y^{j\ell} = D_Y(y_j, y_\ell)$, the squared GW distance is defined as
\begin{equation}
\begin{aligned}
\operatorname{GW}^2(\hat{\mu}, \hat{\nu})
&= \min_{\mathbf{T} \in \Pi(\hat{\mu}, \hat{\nu})}
\sum_{i,j,k,\ell}
\left( D_X^{ik} - D_Y^{j\ell} \right)^2 \,
T_{i,j} \, T_{k,\ell},
\end{aligned}
\label{gromov_wasserstein}
\end{equation}
where $\mathbf{T}$ is a transport plan, $\mathbf{T}^\star$ denotes an optimal transport plan, and $\Pi(\hat{\mu}, \hat{\nu})$ denotes the set of admissible couplings with marginals $\hat{\mu}$ and $\hat{\nu}$.

The semi-relaxed variant (srGW) relaxes one marginal constraint, allowing more flexibility in matching structures \cite{clark2024generalized, van2024distributional}:
\begin{equation}
\begin{aligned}
\operatorname{srGW}^2(\hat{\mu}, \hat{\nu})
&= \min_{\mathbf{T} \in \Pi^{\mathrm{sr}}(\hat{\mu})}
\sum_{i,j,k,\ell}
\left( D_X^{ik} - D_Y^{j\ell} \right)^2 \,
T_{i,j} \, T_{k,\ell},
\end{aligned}
\label{eq:sr_gw}
\end{equation}

Relational structures can be aggregated via the GW barycenter, defined as
\begin{equation}
\bar{D} = \arg\min_{D \in \mathbb{R}^{n \times n}}
\sum_{s=1}^{S} \lambda_s \,
\mathrm{GW}^2\!\left(D^{(s)}, D\right),
\end{equation}
where $\lambda_s \geq 0$ and $\sum_{s=1}^{S} \lambda_s = 1$. In this work, we use uniform weights $\lambda_s = 1/S$ \cite{peyre2016gromov}.

\subsection{GW-MDS Embedding}

Specifically, we seek an embedding $Y \in \mathbb{R}^{n \times d}$ whose induced distance matrix $D_Y$ best matches the relational structure encoded in $D$ \cite{Eufrazio}:
\begin{equation}
\min_{Y \in \mathbb{R}^{n \times d}}
\mathrm{GW}^2\!\left(D, D_Y\right),
\qquad
D_Y(i,j) = \|y_i - y_j\|_2.
\end{equation}

Here, $D$ and $D_Y$ denote pairwise distance matrices associated with empirical measures supported on $\mathcal{X}$ and $\mathcal{Y}$, respectively.

This formulation can be interpreted as a generalized multidimensional scaling problem, where relational alignment is mediated by an optimal transport plan $\mathbf{T}$. 

More explicitly, the objective depends jointly on the embedding $Y$ and the transport plan $\mathbf{T}$, leading to a non-convex optimization problem due to their coupling, which is addressed via an alternating optimization strategy.

The optimization is performed directly over the embedding coordinates $Y$ using gradient-based updates to minimize $\mathrm{GW}^2(D, D_Y)$. In particular, the embedding is iteratively updated as
\begin{equation}
    Y^{(t+1)} = Y^{(t)} - \eta \nabla_Y \mathrm{GW}^2\bigl(D, D_Y^{(t)}\bigr),
\end{equation}
while the transport plan $\mathbf{T}$ is recomputed at each iteration. This alternating procedure jointly estimates the embedding and the relational correspondence between structures.

Importantly, no explicit normalization is applied to $D_Y$, as the scale of the embedding is implicitly determined by the GW objective.

\section{Proposed Multi-View Framework}

In this section, we present a Gromov-Wasserstein-based framework for multi-view data analysis. We first introduce the Bary-GWMDS approach for embedding, followed by the Mean-GWMDS-C method for clustering.

\subsection{Bary-GWMDS}

We introduce Bary-GWMDS (Barycentric Gromov-Wasserstein Multidimensional Scaling) (Algorithm~\ref{alg:bary-gwmds}), a two-stage method for multi-view embedding. The proposed approach first aggregates view-dependent relational structures using a Gromov-Wasserstein (GW) barycenter \cite{peyre2016gromov}, and then computes a low-dimensional embedding by applying Gromov-Wasserstein Multidimensional Scaling (GW-MDS) \cite{Eufrazio} to the resulting consensus metric.

More specifically, given a set of distance matrices $\{D^{(s)}\}_{s=1}^S$, the GW barycenter is used to compute a consensus distance matrix $\bar{D}$ that captures the shared relational structure across views. Subsequently, a low-dimensional embedding $Y$ is obtained by solving a GW-MDS problem based on $\bar{D}$.

\begin{algorithm}[t]
\caption{Bary-GWMDS}
\label{alg:bary-gwmds}
\textbf{Input:} $\{D^{(s)}\}_{s=1}^{S}$, embedding dimension $d$ \\
\textbf{Output:} $Y \in \mathbb{R}^{n \times d}$
\begin{algorithmic}[1]
\State Compute the GW barycenter $\bar{D}$ of $\{D^{(s)}\}_{s=1}^{S}$
\State Initialize $Y \in \mathbb{R}^{n \times d}$
\State Optimize $Y$ by minimizing $\mathrm{GW}(\bar{D}, D_Y)$
\State \textbf{return} $Y$
\end{algorithmic}
\end{algorithm}

\subsection{Mean-GWMDS-C}
Beyond full-support embedding, we consider a clustering-oriented multi-view formulation, termed Mean-GWMDS-C (Mean Gromov-Wasserstein Multidimensional Scaling for Clustering) (Algorithm~\ref{alg:Mean-GWMDS-C}), which aims at learning compact representations defined over a reduced set of representative points. In this setting, the latent embedding $Y \in \mathbb{R}^{k \times d}$, with $k \ll n$, encodes the geometry of learned prototypes rather than individual samples. 

Mean-GWMDS-C operates by first aggregating the view-dependent relational structures through direct averaging of the corresponding distance matrices. Given the averaged multi-view distance matrix, a reduced-support relational representation is learned using a  (semi-relaxed) Gromov-Wasserstein-based \cite{kerdoncuff2021sampled, clark2024generalized}  objective, and the prototype embedding $Y$ is subsequently optimized so as to induce a distance matrix $D_Y$ consistent with the aggregated relational geometry. Solving semi-relaxed Gromov-Wasserstein problems independently across views would induce multiple, view-dependent transport plans, hindering coherent clustering. Mean-GWMDS-C resolves this issue by aggregating relational structures prior to optimization, yielding a single consensus transport plan and favoring structural compactness over pointwise fidelity.

\begin{algorithm}[t]
\caption{Mean-GWMDS-C}
\label{alg:Mean-GWMDS-C}
\textbf{Input:} $\{D^{(s)}\}_{s=1}^{S}$, prototypes $k$, dim $d$ \\
\textbf{Output:} $Y \in \mathbb{R}^{k \times d}$
\begin{algorithmic}[1]
\State $\bar{D} \leftarrow \frac{1}{S}\sum_{s=1}^{S} D^{(s)}$
\State Optimize $Y$ by minimizing $\mathrm{GW}(\bar{D}, D_Y)$
\State \textbf{return} $Y$ \hfill (labels: $\hat{c}(i)=\arg\max_j \mathbf{T}_{ij}$)
\end{algorithmic}
\end{algorithm}

After optimization, the proposed method yields an optimal transport plan \(\mathbf{T} \in \mathbb{R}_+^{n \times K}\). Each entry \(T_{ik}\) quantifies the amount of mass \(\mu_i\) transported from
sample \(i\) to prototype \(k\). Hard cluster assignments are obtained by assigning each sample to the prototype
receiving the largest fraction of its mass,
\(
\hat{c}(i) = \arg\max_{k} T_{ik} 
\) \cite{van2024distributional}.
Soft assignments are given by the normalized transport weights
\(P_{ik} = T_{ik} / \mu_i\).
The total mass accumulated by each cluster,
\(\sum_i T_{ik} \approx \nu_k\), reflects its effective size.
Overall, clustering naturally emerges from optimal mass redistribution between
relational structures, without relying on explicit centroids or decision
boundaries.

As such, Mean-GWMDS-C provides a complementary perspective to Bary-GWMDS, targeting scenarios in which reduced representations and clustering-oriented analysis are preferred over full-support embeddings.

\section{Experiments and Discussion}
In this section, we evaluate the proposed framework through experiments, considering both embedding quality and clustering performance.

\subsection{Bary-GWMDS with Geodesic Distances}
We begin by evaluating the proposed Bary-GWMDS framework. In all experiments, each view is represented by a pairwise distance matrix computed using geodesic distances, which better capture the intrinsic geometry of nonlinear manifolds. Geodesic distances are obtained by constructing a neighborhood graph and computing shortest-path distances.

Preliminary experiments indicate that using Euclidean distances in the barycentric aggregation leads to unstable or distorted consensus structures. In particular, Euclidean metrics tend to overemphasize local linearity and fail to preserve global manifold relationships, which adversely affects the Gromov-Wasserstein barycenter computation.

By contrast, geodesic distances yield well-structured barycenters that preserve the intrinsic relational organization common to all views. This choice results in stable consensus distance matrices and enables the subsequent GW-MDS optimization to produce meaningful low-dimensional embeddings.

To assess the effectiveness of the Bary-GWMDS approach, we compare it with Multi-ISOMAP \cite{rodosthenous2024multi}, a multi-view method based on geodesic distances. Multi-ISOMAP extends the classical ISOMAP framework to the multi-view setting. This baseline is particularly relevant to our study, as both methods rely on geodesic distances to capture the intrinsic geometry of nonlinear manifolds.

\subsection{Multi-View Construction}

To evaluate the proposed methods under controlled geometric variability, we consider synthetic multi-view datasets generated from a common underlying manifold. 

The first view is obtained by applying a rigid rotation in the, which preserves intrinsic manifold distances while altering the extrinsic orientation. The second is generated via a linear deformation involving scaling and shearing, resulting in significant distortion of distances while preserving the topological structure of the manifold. These transformations are designed to simulate realistic multi-view scenarios in which different observations share the same latent geometry but exhibit heterogeneous ambient-space representations.

\begin{figure}[t]
    \centering
    \includegraphics[width=0.48\columnwidth]{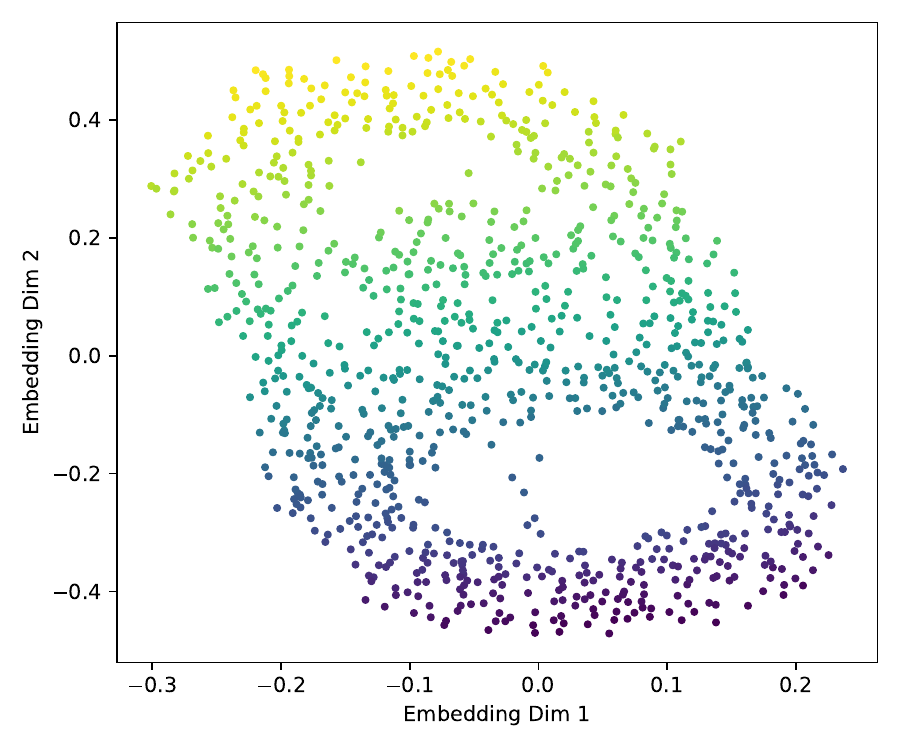 }
    \hfill
    \includegraphics[width=0.48\columnwidth]{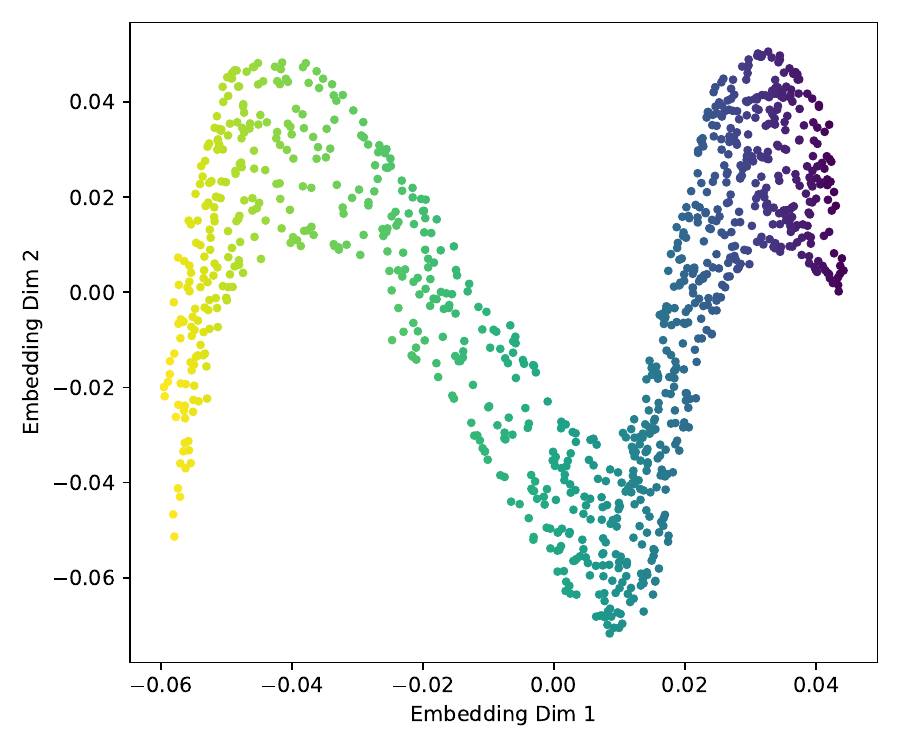}
    \caption{Swiss roll embeddings:
\textbf{Left:} Bary-GWMDS embedding.
\textbf{Right:} Multi-ISOMAP embedding.}
    \label{fig:swissroll_comparison}
\end{figure}

\begin{figure} [t]
    \centering
    \includegraphics[width= 1.0\linewidth]{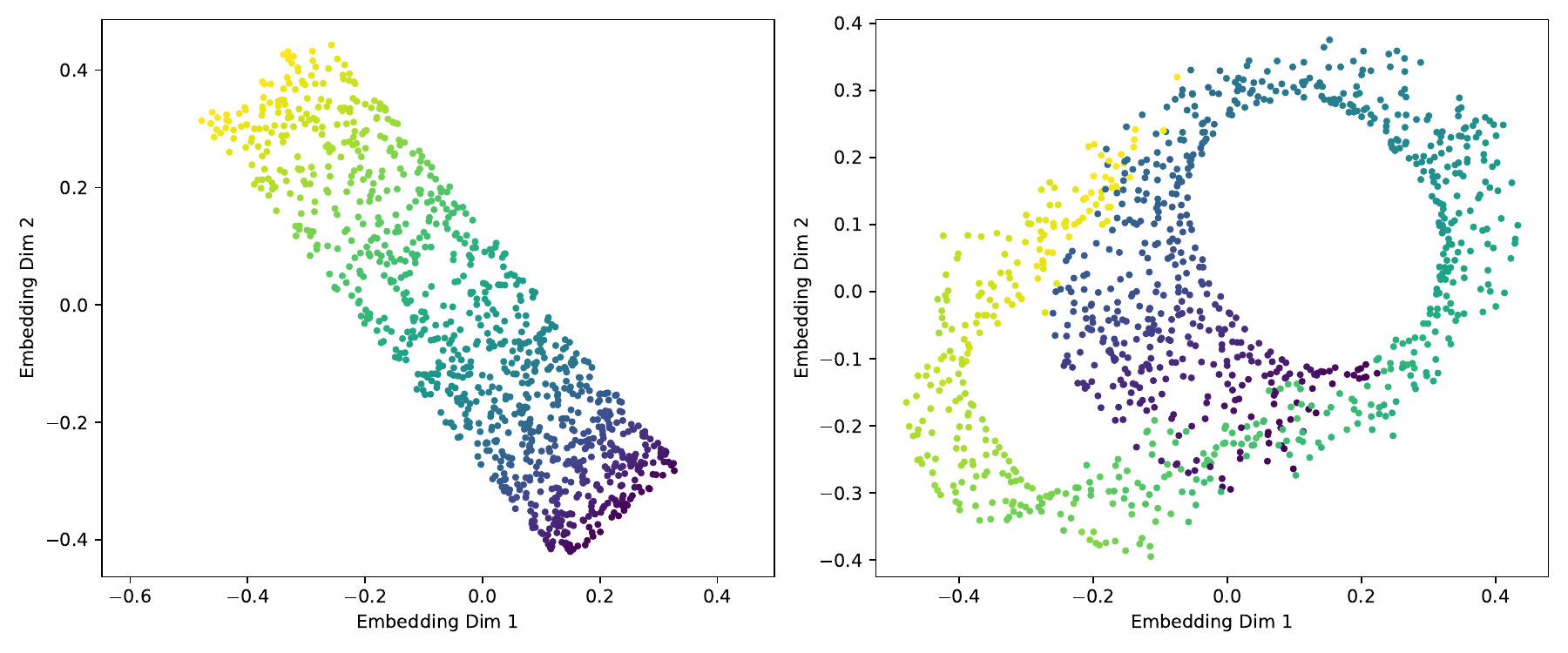}
    \caption{Embeddings obtained separately from each view of the Swiss roll dataset.
\textbf{Left:} embedding View 1.
\textbf{Right:} embedding View 2.}
    \label{fig:swissroll_views}
\end{figure}

\begin{table}[h]
 \centering
    \begin{tabular}{lccccc}
        \toprule
                  & & S-curve & Swiss Roll & Mobius  \\
        \midrule
      Bary-GWMDS  & view 1  & \textbf{0.98} & \textbf{0.96} & \textbf{0.93} \\
               & view 2 & \textbf{0.91} &  0.26 & \textbf{0.96}  \\
               & mean & \textbf{0.945} & 0.610 & \textbf{0.945} \\
      \midrule
      Multi-Isomap  & view 1 & 0.80 & 0.74 & \textbf{0.93}   \\
               & view 2 & 0.85 & \textbf{0.67} & 0.95  \\
               & mean & 0.825 & \textbf{0.705} & 0.940  \\
        \bottomrule
    \end{tabular}
    \caption{View-wise and mean correlation between learned embeddings. }
    \label{tab:correlation_multiview_geodesic}
\end{table}

Table~\ref{tab:correlation_multiview_geodesic} reports view-wise and mean Pearson correlation \cite{benesty2009pearson} between the learned embeddings and the original geodesic distance matrices on synthetic manifolds.
Fig.~\ref{fig:swissroll_views} illustrates the Swiss roll embeddings obtained from each view. For Bary-GWMDS, correlations differ substantially across views
($0.96$ for View~1 and $0.26$ for View~2), reflecting its focus on preserving view-specific relational structure rather than enforcing uniform alignment.

Multi-ISOMAP yields more balanced correlations ($0.74$ and $0.67$), but fails to preserve the intrinsic Swiss roll geometry due to direct distance averaging, which blurs view-dependent relational information. Overall, these results show
that correlation alone does not guarantee geometric fidelity, and that barycentric optimal transport enables more faithful manifold preservation.

\subsection{Mean-GWMDS-C: Clustered Multi-View Representations}

We now consider a clustering-oriented formulation of the proposed framework, termed Mean-GWMDS-C, which learns compact multi-view representations by operating on a reduced set of prototypes. The method aggregates view-wise geodesic distance matrices through direct averaging and learns a reduced-support representation using a \emph{semi-relaxed} Gromov-Wasserstein objective.

In this setting, cluster assignments are obtained from the transport plan
mapping samples to prototypes. We evaluate Mean-GWMDS-C on several real-world multi-view datasets, including MultiFeature (MulFe) \cite{van1998handwritten}, MSRC \cite{shotton2009textonboost}, Handwritten (Han) \cite{chin2022improving}, and Caltech101-7 (Calt) \cite{chen2022low}. The proposed method is compared with Multi-View K-Means \cite{bickel2004multi} and Multi-View Spectral Clustering \cite{kumar2011co}, using the same number of clusters and external evaluation criteria.

\subsubsection{Evaluation Metrics}

Clustering performance is assessed using Normalized Mutual Information (NMI) \cite{strehl2002cluster} and Adjusted Rand Index (ARI) \cite{gao2023overview}, with higher values indicating better agreement with ground-truth labels.

\begin{table}[h]
 \centering
    \begin{tabular}{lccccc}
        \toprule
                  & & MulFe & Calt & Han & MSRC\\
        \midrule
      Mean-GWMDS-C  & NMI  & \textbf{0.942} &  \textbf{0.260} & 0.773 & 0.463\\
               & ARI & \textbf{0.958} & 0.165 & 0.730 & 0.341\\
                
      \midrule
      MV KMeans  & NMI & 0.759 &  0.240 & 0.549 & 0.264\\
               & ARI & 0.773 &  \textbf{0.185} & 0.382 & 0.129\\

               \midrule
       MV Spectral Clust.  & NMI  & 0.933 &  0.249 & \textbf{0.851} & \textbf{0.577}\\
               & ARI & 0.946 &  0.174 & \textbf{0.793} & \textbf{0.495} \\
                     
        \bottomrule
    \end{tabular}
    \caption{Clustering performance comparison in terms of NMI and ARI.}
    \label{tab:Metrica}
\end{table}

Table~\ref{tab:Metrica} reports NMI and ARI clustering results on four multi-view datasets with distinct geometric and semantic characteristics. On the Multiple Features dataset, which exhibits a strong shared
geometry across views, Mean-GWMDS-C achieves the best performance, attaining the highest NMI (0.942) and ARI (0.958). A similar behavior is observed on Caltech101-7, where Mean-GWMDS-C obtains the highest NMI, indicating
the effectiveness of geometry-aware relational aggregation under heterogeneous visual descriptors.

For the Handwritten and MSRC-v5 datasets, Multi-view Spectral Clustering yields the highest scores, benefiting from compact clusters and strong local neighborhood consistency. Nevertheless, Mean-GWMDS-C remains
competitive and consistently outperforms multi-view K-Means.

Overall, these results highlight that Gromov-Wasserstein-based relational aggregation is particularly advantageous when views share compatible geometric structures, positioning Mean-GWMDS-C as a robust alternative for multi-view clustering.

\begin{figure}[t]
    \centering
    \begin{minipage}{0.9\linewidth}
        \centering
        \includegraphics[width=\linewidth]{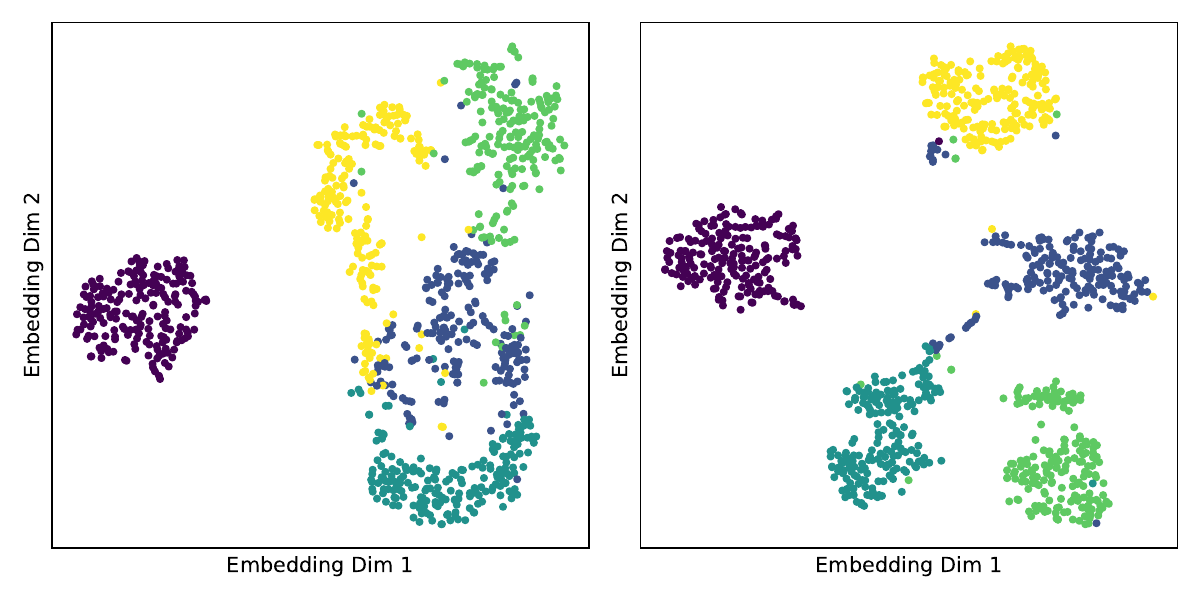}
        \caption*{(a) Two-dimensional t-SNE embeddings of two feature views from the Multiple Features dataset.}
    \end{minipage}
    \begin{minipage}{0.9\linewidth}
        \centering
        \includegraphics[width=\linewidth]{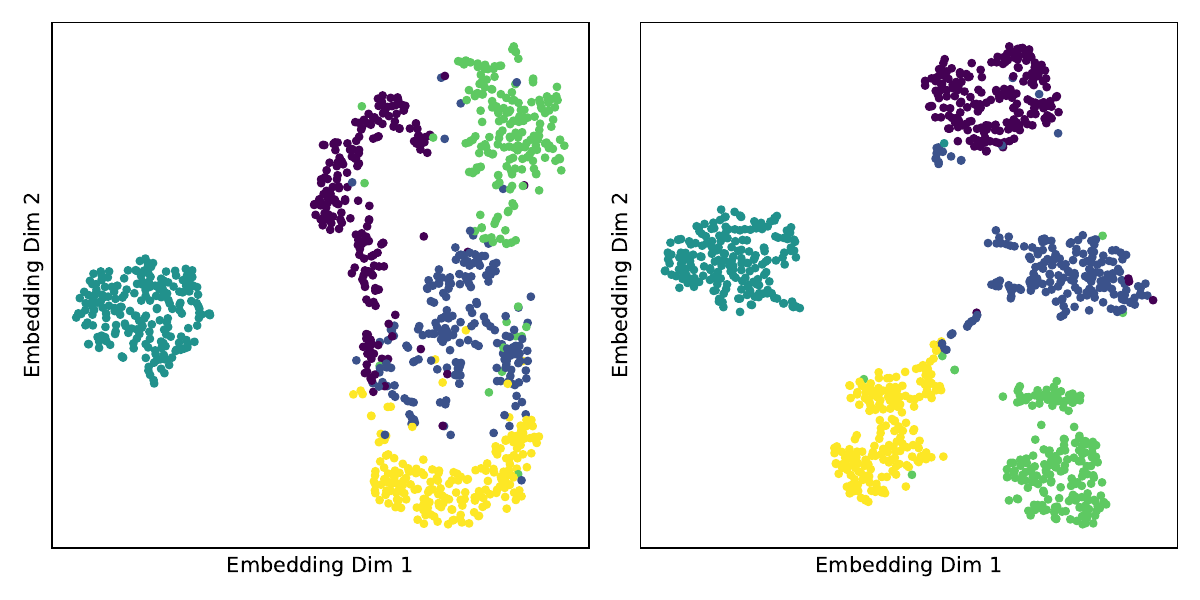}
        \caption*{(b) Visualizations of two views from the Multiple Features dataset, colored by Mean-GWMDS-C clusters.}
    \end{minipage}
    \caption{Visualization of the Multiple Features dataset, showing two feature views and the cluster structure recovered by Mean-GWMDS-C.}
    \label{fig:qualitative_mean-gwmds-c}
\end{figure}

Fig. \ref{fig:qualitative_mean-gwmds-c} provides a qualitative illustration of
the clustering behavior of Mean-GWMDS-C on the Multiple Features dataset. This visualization highlights the ability of Mean-GWMDS-C to integrate
multi-view relational information into a compact representation while
preserving meaningful geometric separation between clusters.

\begin{figure}[t]
\centering

\begin{minipage}{0.48\linewidth}
    \centering
    \includegraphics[width=\linewidth]{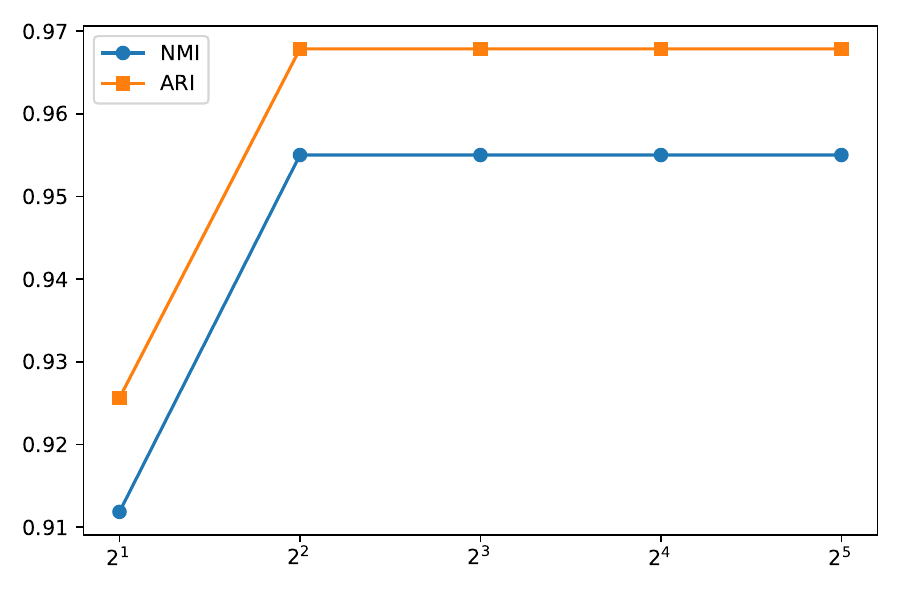}
    \caption*{(a)  MultiFeature}
\end{minipage}\hfill
\begin{minipage}{0.48\linewidth}
    \centering
    \includegraphics[width=\linewidth]{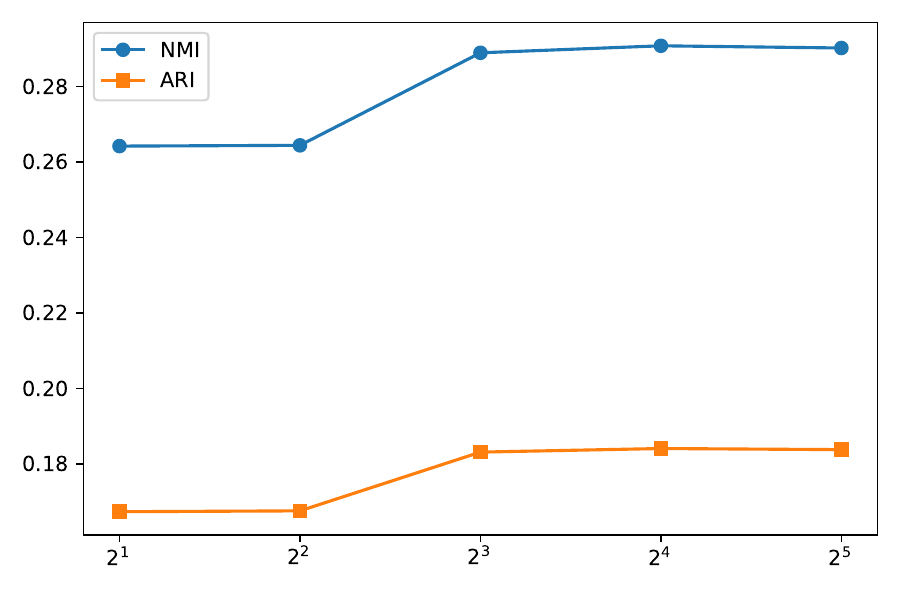}
    \caption*{(b) Caltech101-7}
\end{minipage}

\vspace{2mm}

\begin{minipage}{0.48\linewidth}
    \centering
    \includegraphics[width=\linewidth]{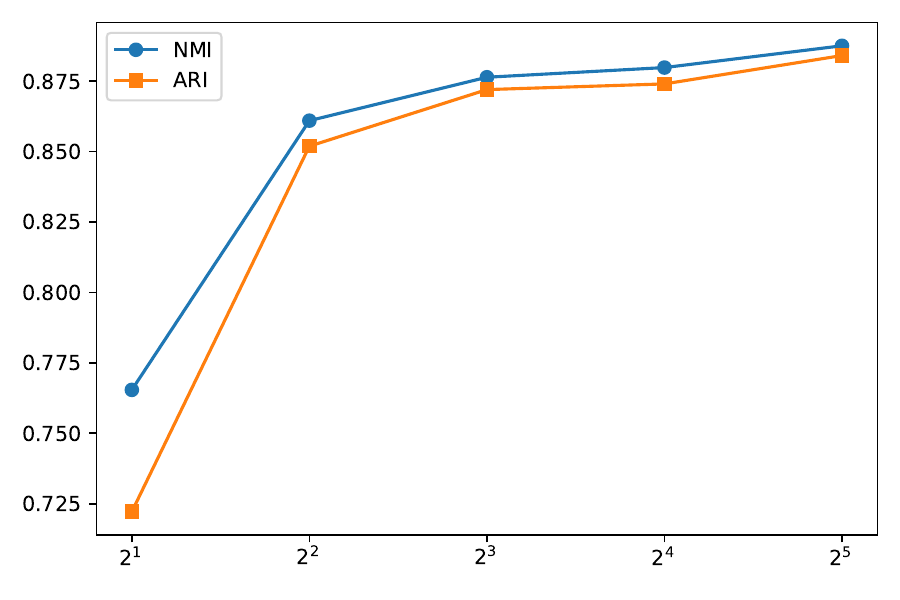}
    \caption*{(c) Handwritten}
\end{minipage}\hfill
\begin{minipage}{0.48\linewidth}
    \centering
    \includegraphics[width=\linewidth]{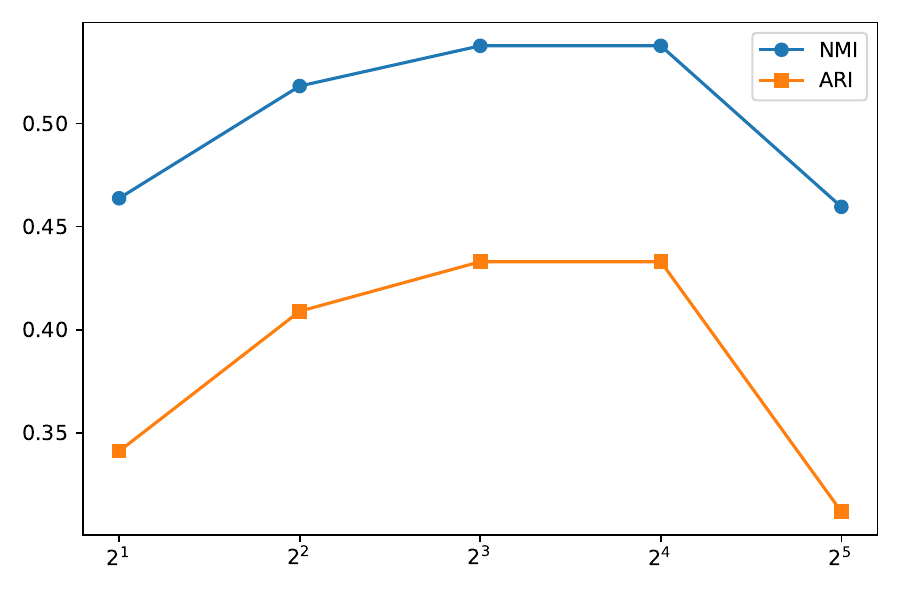}
    \caption*{(d) MSRC}
\end{minipage}

\caption{Clustering performance of Mean-GWMDS-C as a function of the embedding dimensionality on different multi-view datasets, evaluated using NMI and ARI. }
\label{fig:multifeature_2x2}
\end{figure}

\begin{table}[t]
\centering
\setlength{\tabcolsep}{3pt}
\begin{tabular}{c|cc|cc|cc|cc}
\toprule
 & \multicolumn{8}{c}{\textbf{NMI / ARI}} \\
$\dim Y$ 
& \multicolumn{2}{c}{\textsc{MulFe}}
& \multicolumn{2}{c}{\textsc{Calt}}
& \multicolumn{2}{c}{\textsc{Han}}
& \multicolumn{2}{c}{\textsc{MSRC}} \\
\cmidrule(lr){2-3} \cmidrule(lr){4-5}
\cmidrule(lr){6-7} \cmidrule(lr){8-9}
 & NMI & ARI & NMI & ARI & NMI & ARI & NMI & ARI \\
\midrule
2  & 0.9118 & 0.9256 & 0.2642 & 0.1673 & 0.7654 & 0.7221 & 0.4638 & 0.3411 \\
4  & 0.9550 & 0.9679 & 0.2644 & 0.1675 & 0.8610 & 0.8519 & 0.5182 & 0.4090 \\
8  & 0.9550 & 0.9679 & 0.2890 & 0.1831 & 0.8764 & 0.8720 & 0.5377 & 0.4330 \\
16 & 0.9550 & 0.9679 & 0.2908 & 0.1841 & 0.8798 & 0.8740 & 0.5377 & 0.4330 \\
32 & 0.9550 & 0.9679 & 0.2902 & 0.1838 & 0.8875 & 0.8841 & 0.4596 & 0.3117 \\
\bottomrule
\end{tabular}
\caption{Clustering performance of Mean-GWMDS-C as a function of the embedding dimensionality.}
\label{tab:dim_sensitivity}
\end{table}

Table \ref{tab:dim_sensitivity} and Fig. \ref{fig:multifeature_2x2} show the impact of the embedding dimensionality on Mean-GWMDS-C. The method achieves stable performance for low-dimensional embeddings on  MultiFeatur and Caltech101-7, while moderate increases in dimension benefit the Handwritten and MSRC datasets. Overall, the results indicate that Mean-GWMDS-C is robust to the choice of embedding dimension and performs
well with compact latent representations.

\begin{figure}[t]
\centering
\begin{minipage}{0.48\linewidth}
    \centering
    \includegraphics[width=\linewidth]{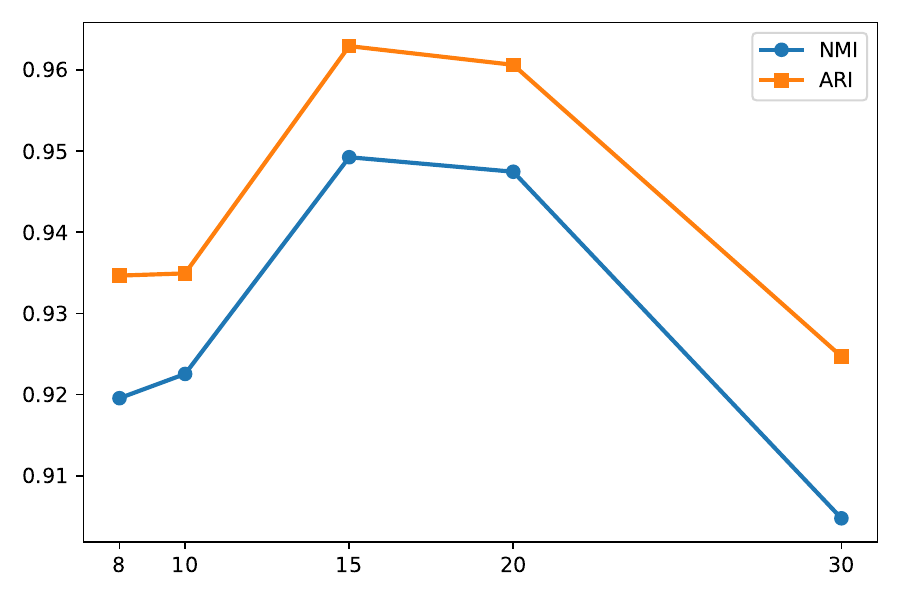}
    \caption*{(a)  MultiFeature}
\end{minipage}\hfill
\begin{minipage}{0.48\linewidth}
    \centering
    \includegraphics[width=\linewidth]{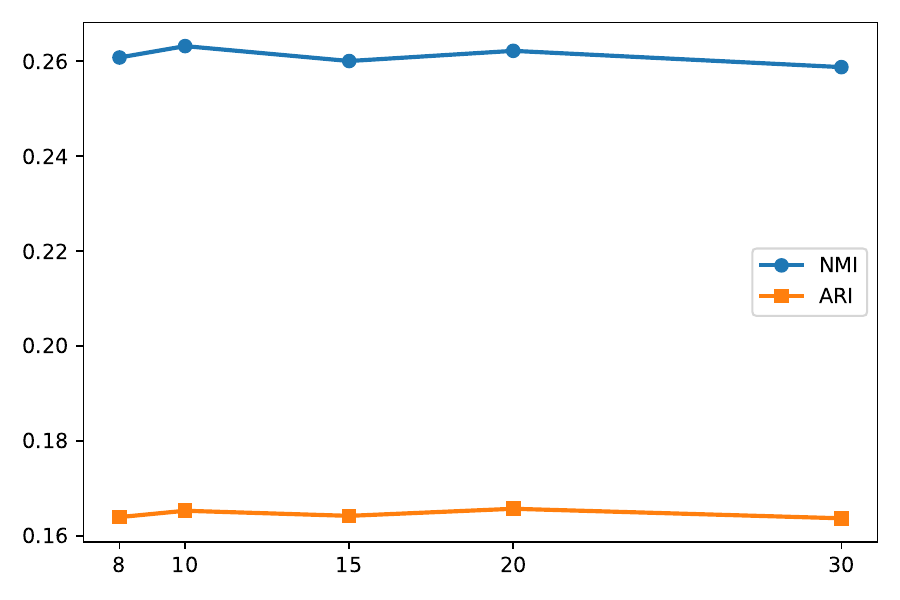}
    \caption*{(b) Caltech101-7}
\end{minipage}

\vspace{2mm}

\begin{minipage}{0.48\linewidth}
    \centering
    \includegraphics[width=\linewidth]{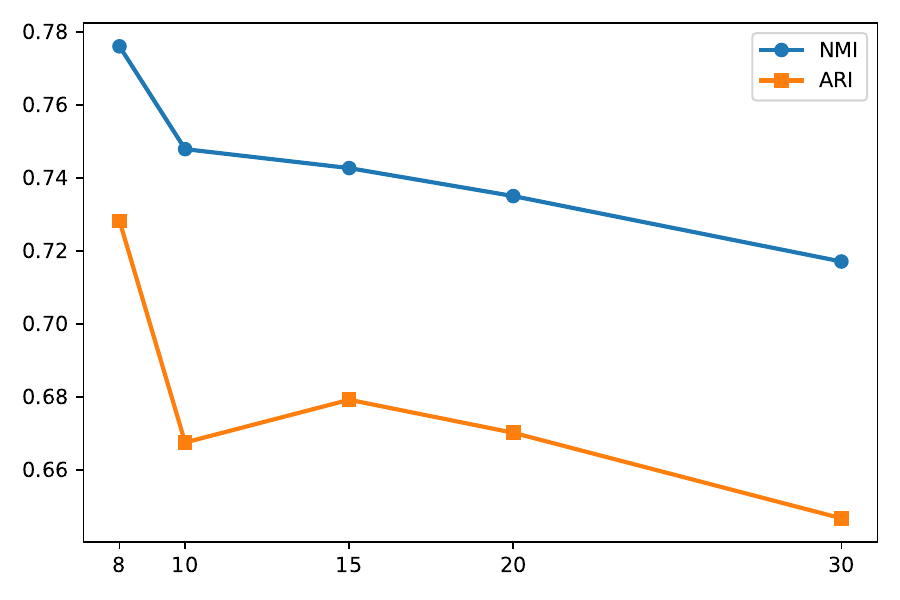 }
    \caption*{(c) Handwritten}
\end{minipage}\hfill
\begin{minipage}{0.48\linewidth}
    \centering
    \includegraphics[width=\linewidth]{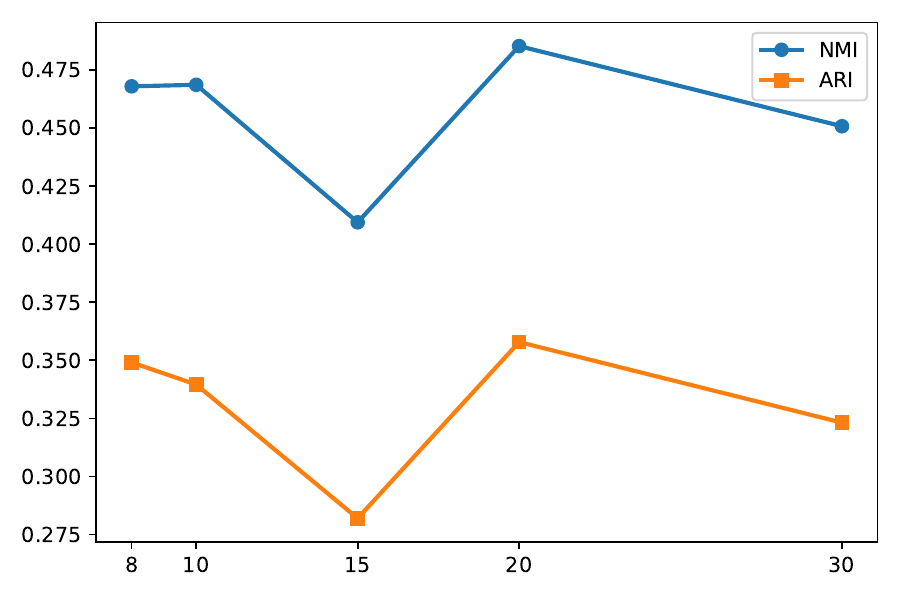 }
    \caption*{(d) MSRC}
\end{minipage}
\caption{Comparison of NMI and ARI as a function of the number of neighbors on different multi-view datasets.}
\label{fig:multifeature_2x2 vizinhos}
\end{figure}

Fig.~\ref{fig:multifeature_2x2 vizinhos} illustrates the sensitivity of Mean-GWMDS-C to the number of neighbors used in the construction of relational graphs. Across all datasets, clustering performance remains relatively stable within a moderate range of neighborhood sizes, indicating robustness to this hyperparameter. On datasets such as MulFe and Handwritten, NMI and ARI exhibit a clear peak for intermediate neighborhood values, while overly small or large neighborhoods tend to degrade performance. Overall, these results suggest that Mean-GWMDS-C benefits from a balanced neighborhood selection, avoiding both overly local and excessively dense relational structures.


\section{Conclusion}

The results presented in this work show that explicitly modeling relational geometry is crucial for multi-view representation learning. By relying on Gromov-Wasserstein alignment, Bary-GWMDS preserves intrinsic geometric
structures that are often degraded by direct distance averaging. In addition, the clustering-oriented Mean-GWMDS-C demonstrates that compact representations can be learned while maintaining meaningful relational consistency across
views. These findings highlight the importance of geometry-aware optimal transport formulations for both embedding and clustering in multi-view settings.


\bibliographystyle{IEEEtran}

\bibliography{sn-bibliography}







\end{document}